\documentclass[10pt,twocolumn,letterpaper]{article}
\pdfoutput=1
\usepackage{arydshln}
\usepackage{multirow}
\usepackage{colortbl}
\usepackage{subfigure}
\usepackage{amssymb}
\usepackage{bbding}
\usepackage[pagenumbers]{iccv}      

\definecolor{iccvblue}{rgb}{0.21,0.49,0.74}
\usepackage[pagebackref,breaklinks,colorlinks,allcolors=iccvblue]{hyperref}

\def\paperID{9266} 
\def\confName{ICCV}
\def\confYear{2025}

\title{ActionArt: Advancing Multimodal Large Models for Fine-Grained Human-Centric Video Understanding}
  
\author{Yi-Xing Peng$^{1,2,3,4}$, Qize Yang$^{2}$, Yu-Ming Tang$^{1,4}$, Shenghao Fu$^{1,4}$, \\ Kun-Yu Lin$^{1}$,  Xihan Wei$^{2}$, Wei-Shi Zheng$^{1,3,4}$\thanks{: Corresponding author. Dataset is available in \url{https://www.modelscope.cn/datasets/maybex/ActionArt} } \\
$^1$ Sun Yat-sen University, China; \qquad$^2$ Tongyi Lab, Alibaba Group; \\
$^3$Peng Cheng Laboratory, Shenzhen, China;
\\ $^4$Key Laboratory of Machine Intelligence and Advanced Computing, Ministry of Education, China. 
\\
{\tt\small {pengyx23}@mail2.sysu.edu.cn}
{\tt\small {qize.yqz}@alibaba-inc.com}
{\tt\small {wszheng}@ieee.org}
\\
}

\begin{document}
\maketitle
\begin{abstract}
Fine-grained understanding of human actions and poses in videos is essential for human-centric AI applications. In this work, we introduce \textbf{ActionArt}, a fine-grained video-caption dataset designed to advance research in human-centric multimodal understanding. Our dataset comprises thousands of videos capturing a broad spectrum of human actions, human-object interactions, and diverse scenarios, each accompanied by detailed annotations that meticulously label every limb movement.
We develop eight sub-tasks to evaluate the fine-grained understanding capabilities of existing large multimodal models across different dimensions. Experimental results indicate that, while current large multimodal models perform commendably on various tasks, they often fall short in achieving fine-grained understanding. We attribute this limitation to the scarcity of meticulously annotated data, which is both costly and difficult to scale manually.
Since manual annotations are costly and hard to scale, we propose proxy tasks to enhance the model perception ability in both spatial and temporal dimensions. These proxy tasks are carefully crafted to be driven by data automatically generated from existing MLLMs, thereby reducing the reliance on costly manual labels. Experimental results show that the proposed proxy tasks significantly narrow the gap toward the performance achieved with manually annotated fine-grained data.

\end{abstract}

\section{Introduction}
\label{sec:intro}
A fine-grained understanding of human actions in videos is vital for advancing various human-centric AI applications, such as enhanced human-AI interactions in VR systems and sophisticated surveillance systems \cite{anet,khirodkar2024sapiens,yang2025omni,zhao2025humanomni}. These applications depend on AI systems' ability to accurately interpret subtle and complex human activities within dynamic environments. Although inherently multimodal \cite{alayrac2022flamingo,zhao2025llavaoctopusunlockinginstructiondrivenadaptive,dai2023instructblip}, the use of multimodal large models for detailed human-centric video understanding is still a relatively untapped area of exploration.

\begin{figure}[tp]
    \centering
    \includegraphics[width=\linewidth]{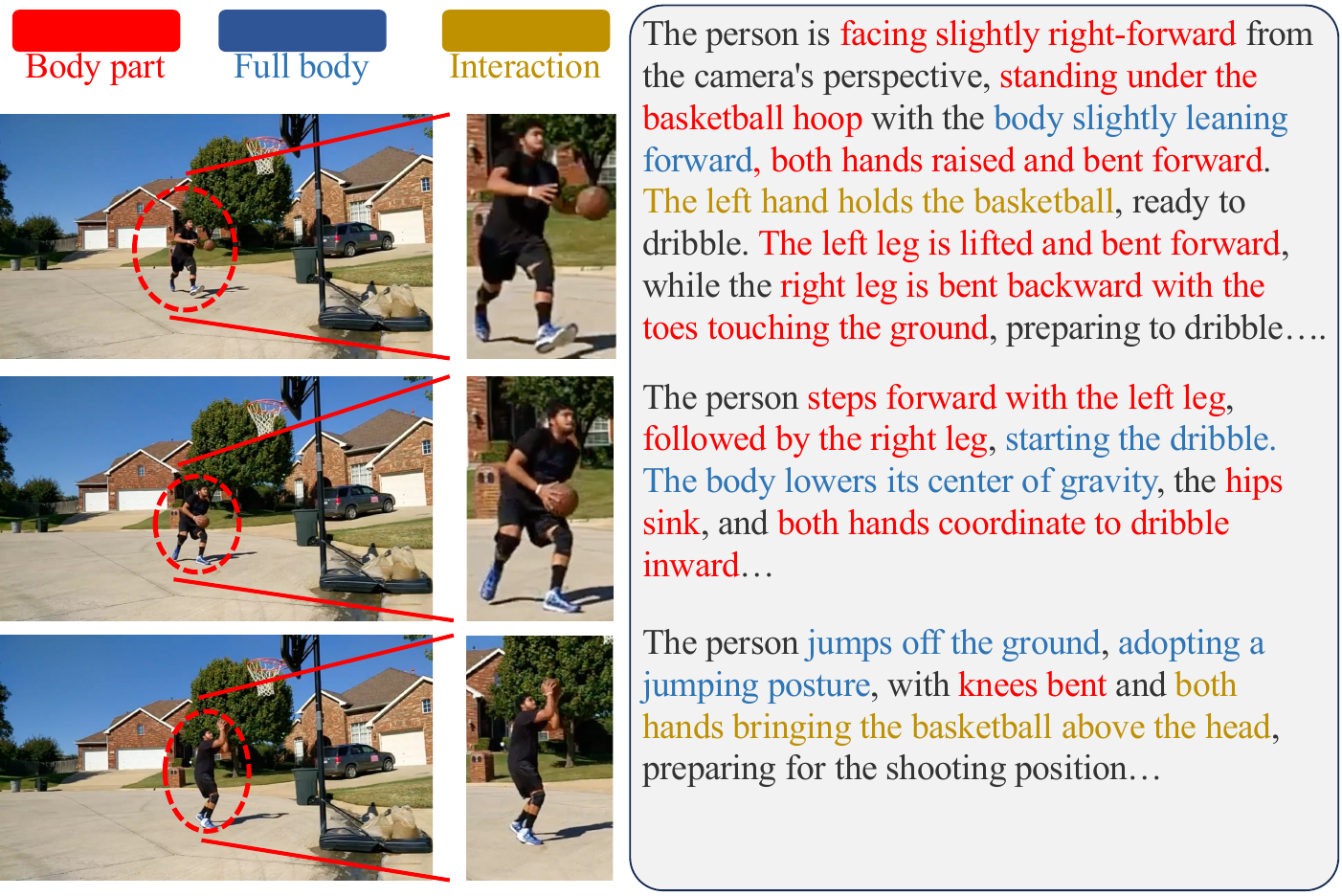}
    \caption{An example of the caption for consecutive frames in our ActionArt dataset. We provide detailed captions that describe human poses, body parts, and interactions with objects. The captions also convey temporal changes in the video at a fine-grained level.}
    \label{fig:intro}
    \vspace{-6pt}
\end{figure}

In this work, we introduce \textbf{ActionArt}, a human-centric video dataset enriched with fine-grained manual captions, aimed at driving research in detailed video understanding. 
Our dataset builds on the MoVid~\cite{motionllm} dataset, which comprises thousands of videos showcasing a wide spectrum of human actions and interactions across diverse scenarios. Since MoVid lacks detailed captions, we meticulously annotated each video with comprehensive descriptions, providing detailed annotations for every limb movement, as illustrated in Fig.~\ref{fig:intro}. Based on these annotations, we present a new benchmark to thoroughly evaluate the fine-grained understanding capabilities of multimodal large language models (MLLM).

Although existing MLLMs excel across various tasks~\cite{qwen2.5,Qwen2-VL,patraucean2023perception,zhou2024mlvu,li2023seed}, our experiments have identified a significant limitation: their ability to discern fine-grained details in human actions is constrained. We mainly attribute this to the lack of finely annotated data available for training these models. Given the high costs associated with producing such detailed annotations~\cite{chen2024sharegpt4video,rawal2024cinepile,mangalam2024egoschema,Maaz2023VideoChatGPT,tarsier,islam2024video}, our work addresses the specific challenges of fine-grained human-centric video understanding by proposing corresponding proxy tasks. These tasks are designed to leverage automatically annotated data, thereby enhancing the fine-grained perception ability of MLLM without the need for manual annotation efforts.

Specifically, we have identified three key capabilities essential for fine-grained human-centric video understanding. First, the model must accurately interpret human poses in each individual frame. This involves recognizing and analyzing the position and orientation of various body parts~\cite{sun2024hicmae,damonlpsg2023videollama,ye2024mplugowl3longimagesequenceunderstanding,xue2024xgenmmblip3familyopen}. To enhance this capability, we employ advanced MLLMs to generate detailed pose descriptions for individual human images, establishing ``pose description'' as one of our proxy tasks.
Second, the model needs to comprehend the changes in human poses between consecutive frames. This temporal understanding enables the model to effectively capture dynamic movements and transitions. To address this, we introduce a proxy task called ``spatial difference mining'', which challenges the model to identify differences between human poses in adjacent frames.

Finally, the model should be capable of understanding the temporal coherence and sequential context of human actions. This involves not only identifying discrete actions but also perceiving the flow and interconnections of these actions over time~\cite{blip3video-xgenmmvid,tong2022videomae,tong2024cambrian,Maaz2024VideoGPT+}, enabling a comprehensive interpretation of human activities within the video. To enhance this capability, we have developed an innovative pipeline for automatically generating video question-answer (QA) pairs that support fine-grained temporal understanding. 
Rather than directly generating QA pairs from the entire video, our approach infers fine-grained information from pairwise frames and synthesizes the long video with corresponding QAs by concatenating these pairwise frames. In this way, we effectively alleviate the noise during the QA generations.

Our main contributions are threefold: (1) we introduce ActionArt, a challenging benchmark focused on fine-grained human-centric video understanding, complete with a comprehensive evaluation protocol. (2) We reveal that existing MLLMs demonstrate unsatisfactory performance on this benchmark. To counter this issue, we dissect the challenges involved and design corresponding proxy tasks, offering an end-to-end solution without requiring manual annotations. (3) Our experiments demonstrate that these proxy tasks substantially improve the fine-grained comprehension capabilities of multimodal models.

\section{Related work}
\label{sec:related}

\subsection{Multimodal Large Language Models}
Multimodal large language models (MLLMs) have demonstrated remarkable performance across a diverse range of applications, including Optical Character Recognition (OCR)~\cite{UReader}, mathematics~\cite{mathvista}, and Visual Question Answering (VQA)~\cite{llava}. Current MLLMs leverage vision-language models such as CLIP~\cite{clip} or SigLIP~\cite{siglip} to extract visual tokens, which are then transformed into the large language model (LLM) feature space via a connector. This methodology facilitates the integration of textual human instructions with visual context. The LLM processes both visual tokens and textual instructions, thereby enabling it to respond to queries based on the visual context.

In the domain of image-based MLLMs, research has predominantly concentrated on overcoming the limitations of fixed image resolution~\cite{llavahd, llavahr,zhang2024mme} and enhancing connectors to optimize the interface between vision-language models and LLMs~\cite{pargo,tong2024cambrian}. LLaVA~\cite{llava} proposes using a simple fully-connected layer as a connector while BLIP-2~\cite{blip2} introduces a Q-Former to extract visual information relevant to the textual input. Honeybee~\cite{honeybee} suggests employing convolution blocks as connectors. DeCo~\cite{deco} utilizes pooling layers for token reduction while maintaining effective MLP blocks, and DC~\cite{dense} unifies visual tokens from multiple layers of the VLM within the connector.

When expanding image MLLMs into the video domain~\cite{damonlpsg2023videollama,damonlpsg2024videollama2,apollo}, existing methods concatenate visual tokens from each frame for video representation~\cite{xu2024pllava,lin2023video,wang2024internvideo2} or employ a Q-former to reduce the token load~\cite{li2024mvbench,jin2023chat}. Xu et al.~\cite{sf_llava} propose organizing visual tokens in a SlowFast~\cite{slowfast} manner. VideoChat2~\cite{li2024mvbench} utilizes the Q-former for visual token compression. Oryx-MLLM~\cite{oryx} introduces a dynamic compressor to reduce visual tokens via varying pooling strides.

\begin{figure*}[tp]
    \centering
    \includegraphics[width=\linewidth]{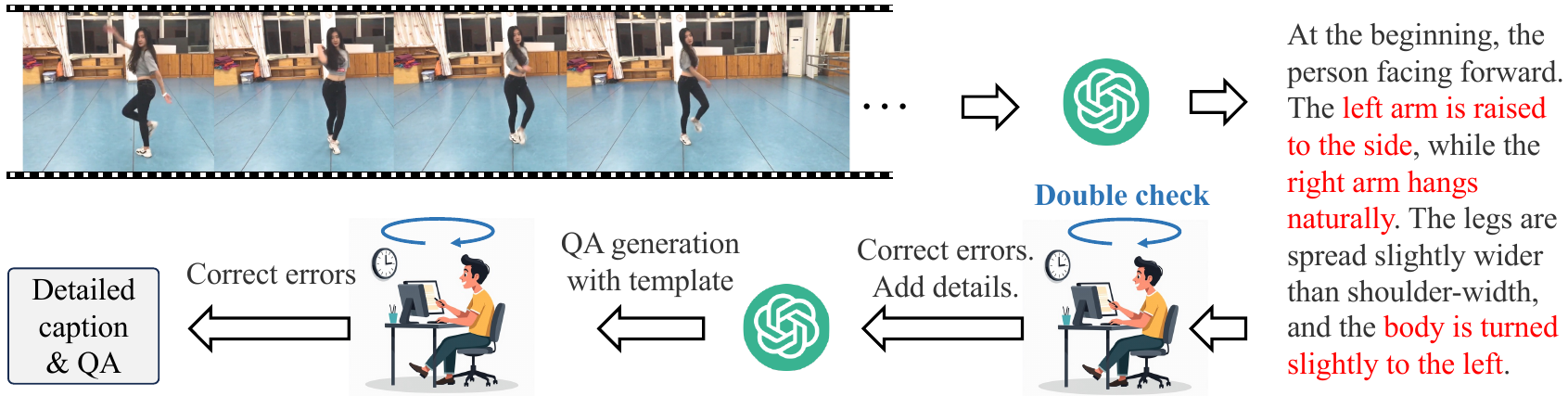}
    \caption{Our annotation pipeline starts by prompting GPT-4V to generate an initial caption for each video. We have observed that existing multimodal large language models (MLLMs) often struggle to accurately discern the directions of human body parts, resulting in coarse-grained descriptions, even with carefully designed prompts. Besides, the caption from GPT is noisy, and {\color{red}the errors} within the captions {\color{red}are highlighted in red.} After identifying these issues, we manually intervene and then prompt GPT-4 to generate various types of QA using corresponding templates, followed by manual refinement.}
    \label{fig:pipeline}
\end{figure*}

\subsection{Benchmarks in Video MLLMs}

Several benchmarks have been introduced to evaluate the capabilities of models in understanding video~\cite{cai2024temporalbenchbenchmarkingfinegrainedtemporal,shangguan2025tomato}. VideoMME~\cite{fu2024video} is designed for a comprehensive assessment of MLLMs' abilities to grasp temporal dimensions, process multimodal inputs such as subtitles and audio, and adapt to complex scenarios. This dataset includes 900 videos sourced from various domains, totaling 254 hours, and features 2,700 human-annotated question-answer pairs. MVBench~\cite{li2024mvbench} focuses on evaluating MLLMs' performance in time-sensitive video understanding tasks, such as action recognition and temporal reasoning in complex scenes. It consists of 20 time-understanding tasks, with videos mainly around 20 seconds in length. Ziyao et al.~\cite{shangguan2025tomato} propose a high-quality video benchmark that eliminates shortcuts in video QA, ensuring that the model must engage in thorough reasoning across the video to arrive at the correct answer. Despite the significant progress in temporal understanding, these works cannot evaluate the fine-grained understanding capabilities of MLLMs.

Concurrently, MoVid~\cite{motionllm} has been developed to assess model performance in the realm of human motion understanding; however, it falls short in providing fine-grained captions that detail human movements within each video. MotionBench~\cite{motionbench} addresses this gap by sourcing data from diverse domains and offering comprehensive captions for each video, which describe human movements in detail.

In contrast, our ActionArt emphasizes human-centric video understanding, with evaluation subtasks that focus on both fine-grained spatial perception and temporal understanding, which distinctly differ from those in MotionBench. Furthermore, while previous efforts primarily aim at enhancing network architecture to process more frames for fine-grained comprehension, our strategy is centered around a data-driven approach. This includes the development of new proxy tasks and the utilization of flexible data collection methods to improve model performance.

\section{ActionArt}
\label{sec:formatting}

Despite significant advancements in human action recognition and video understanding, achieving fine-grained understanding, such as capturing the nuanced movements of body parts and orientations remains underexplored.

To advance research in fine-grained human-centric video understanding, we introduce the ActionArt dataset. Our dataset contains meticulously curated videos displaying diverse human actions, each accompanied by detailed annotations, serving as a robust benchmark for analysis. Furthermore, we have designed seven sub-tasks to comprehensively evaluate the ability of multimodal large language models (MLLMs) in human-centric video understanding.

\subsection{Data}

We utilize the MoVid~\cite{motionllm} dataset as our video source due to its extensive collection of human movement records. Since MoVid does not contain detailed human captions for each video, we have processed a total of 4,009 video clips, each provided with meticulously detailed, fine-grained caption annotations. These videos vary in length and depict a wide range of human activities and human-object interactions across multiple domains, including martial arts, dance, and both indoor and outdoor settings. This diversity presents a significant challenge in understanding human actions within this benchmark.

To capture the intricacies of human movements in the videos, we designed specific prompts and utilized advanced multimodal large models like GPT-4V~\cite{openai2023gpt4v} to generate preliminary captions. Given the current limitations of these models in comprehensively understanding fine-grained human actions, each caption was manually refined to ensure accuracy and completeness. The annotation process is rigorously double-checked to maintain high quality.

\begin{figure}[tp]
    \centering
    \includegraphics[width=\linewidth]{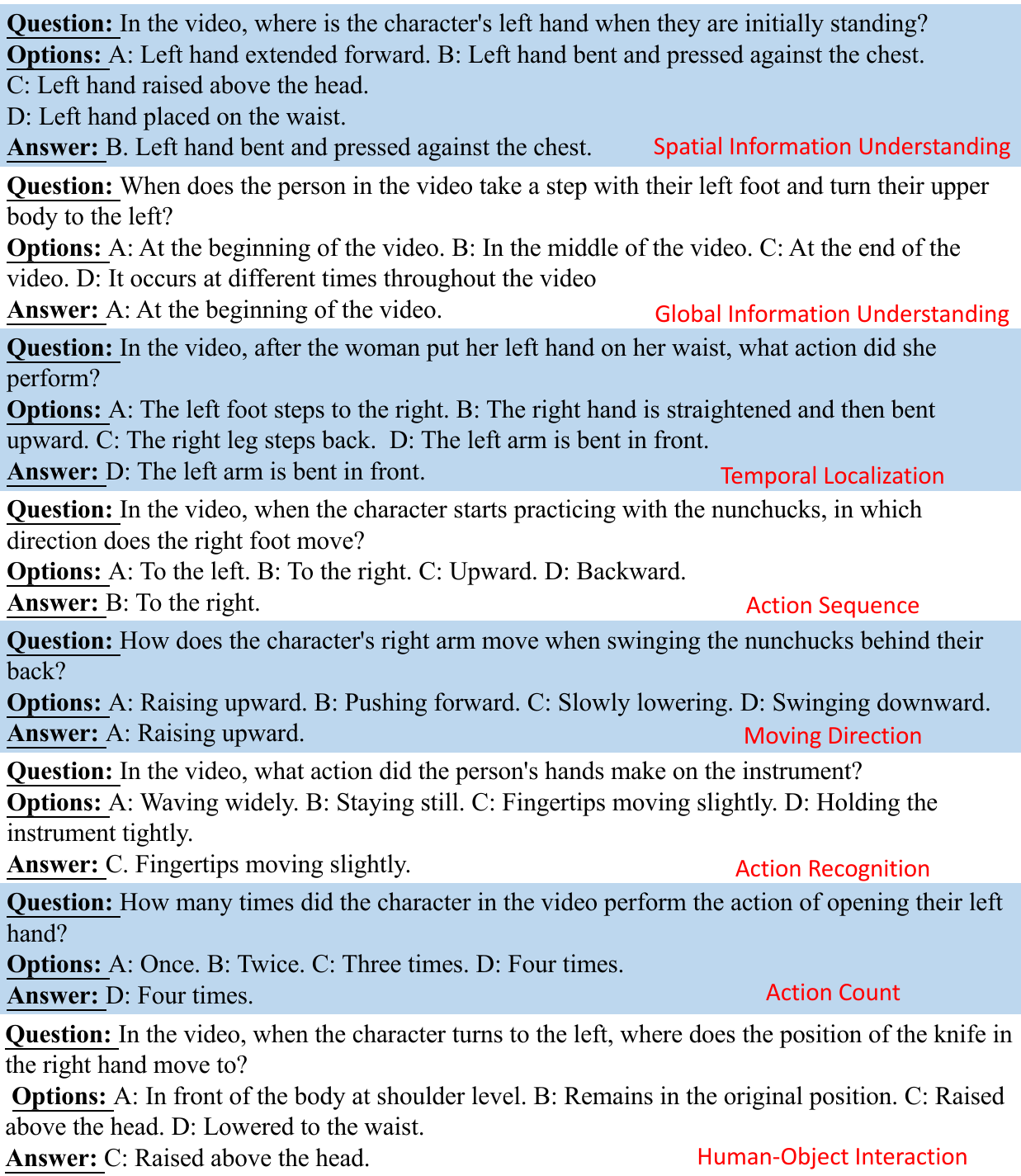}
    \caption{Examples of different types QA.}
    \label{fig:QA}
\end{figure}

\subsection{Benchmark for fine-grained human-centric video understanding}
To thoroughly evaluate the fine-grained understanding capabilities of existing large models, we have designed seven sub-tasks inspired by previous research~\cite{li2024mvbench,motionbench,motionllm} and form a benchmark. These tasks assess models across multiple dimensions, covering global to local and temporal to spatial perspectives. The seven sub-tasks are: local spatial understanding, global spatial understanding, temporal localization, action sequence, moving direction, action understanding, and action count.

To ensure fair and accurate evaluation, we designed the QA in a multiple-choice format as in previous works~\cite{patraucean2023perception,li2024mvbench,motionllm}.
An illustration of each task is shown in Fig.~\ref{fig:QA}

\noindent \textbf{- Local Spatial Understanding (LS).}
This task evaluates the model's ability to discern the posture and position of specific body parts, such as hands and thighs. For instance, determining the placement of a hand or the posture of a foot.

\noindent \textbf{- Global Spatial Understanding (GS).}
Complementing local spatial understanding, this task evaluates a model's capability to recognize the overall posture and position of an individual within a video frame. It involves determining whether a person is standing, lying down, or in another posture, thereby challenging models to integrate spatial cues from the entire frame.

\noindent \textbf{- Temporal Localization (TL).}
This task evaluates a model's ability to accurately pinpoint the timing of specific actions within a video sequence, such as identifying when a person begins to squat. The model must accurately watch the entire video sequence and identify all the fine-grained actions within it to complete the task.

\noindent \textbf{- Action Sequence (AS).}
This task challenges the model to recognize and interpret the sequence of movements involving various body parts, such as the order of arm and leg movements in dance or martial arts. It necessitates the integration of temporal and spatial information.

\noindent \textbf{- Moving Direction (MD).}
This task assesses a model's ability to determine the directional movement of a person or specific body parts. By requiring models to interpret spatial orientation and movement dynamics in three-dimensional space, it emphasizes the integration of spatial and temporal information for comprehensive movement understanding.

\noindent \textbf{- Action Recognition (AR).}
This task focuses on recognizing the specific actions performed by body parts, especially hands and feet, such as waving a hand or tapping a foot. It challenges models to explore detailed movement semantics, advancing nuanced action recognition beyond traditional analysis.

\noindent \textbf{- Action Count (AC).}
This task assesses a model's ability to accurately count the frequency of a specific action within a video, like counting hand waves or squats. It highlights the importance of temporal consistency in modeling human behavior, contributing to robust applications in various fields.

\noindent \textbf{- Human-Object Interaction (HOI).}
In human-centric videos, interactions between people and objects are quite common, such as picking up or discarding an item, or using props for a performance. Consequently, in fine-grained understanding, it is crucial to evaluate a model's comprehension of human-object interactions.

\subsection{Question-Answer Pairs Generation.}
We generated diverse multiple-choice QA pairs for each type of sub-task to measure a model's performance on that sub-task.
For each sub-task, we manually defined the question templates and examples. We fed the manually fine-annotated captions into large language models, such as GPT-4, along with the templates and examples, guiding the large language model to generate the desired QA pairs. 

Despite the presence of manual annotated fine-grained captions, the QA pairs generated from LLM often contained errors, including conflicting options and incorrect answers, which further demonstrates the challenging of fine-grained human-centric video understanding. We manually reviewed the test videos and corrected each question and answer. Ultimately, we obtained 2,678 QA pairs.

\section{Proxy task learning}

We believe that the current models struggle to learn fine-grained action understanding primarily due to a lack of necessary training data. Although precisely labeled data is expensive and difficult to scale, we have fortunately deconstructed the challenges of fine-grained understanding into three components and have designed auxiliary tasks accordingly to enhance the model's capabilities. Through ingenious design, the data for these auxiliary tasks can be generated at scale through automated production processes.

Specifically, to understand the video in fine-grained level, the model should have the three crucial abilities. (1) First, the model is expected to understand the static spatial information in each frame. (2) Second, the model has capture the dynamic change in spatial information between adjacent frames. (3) Moreover, the model should understand the temporal information in the video for action recognition and understanding the action sequence, and so forth.

Based on our analysis, we develop three proxy tasks to effectively utilize the advanced MLLM for automatically generating labeled data for learning, approaching the performance of finetuning with manual labels.
The model is expected to perform well after learning the proxy tasks.

 \begin{figure}[]
   \centering
   \subfigure[Pose description task.]{
        \label{fig:PD}
       \includegraphics[width=\linewidth]{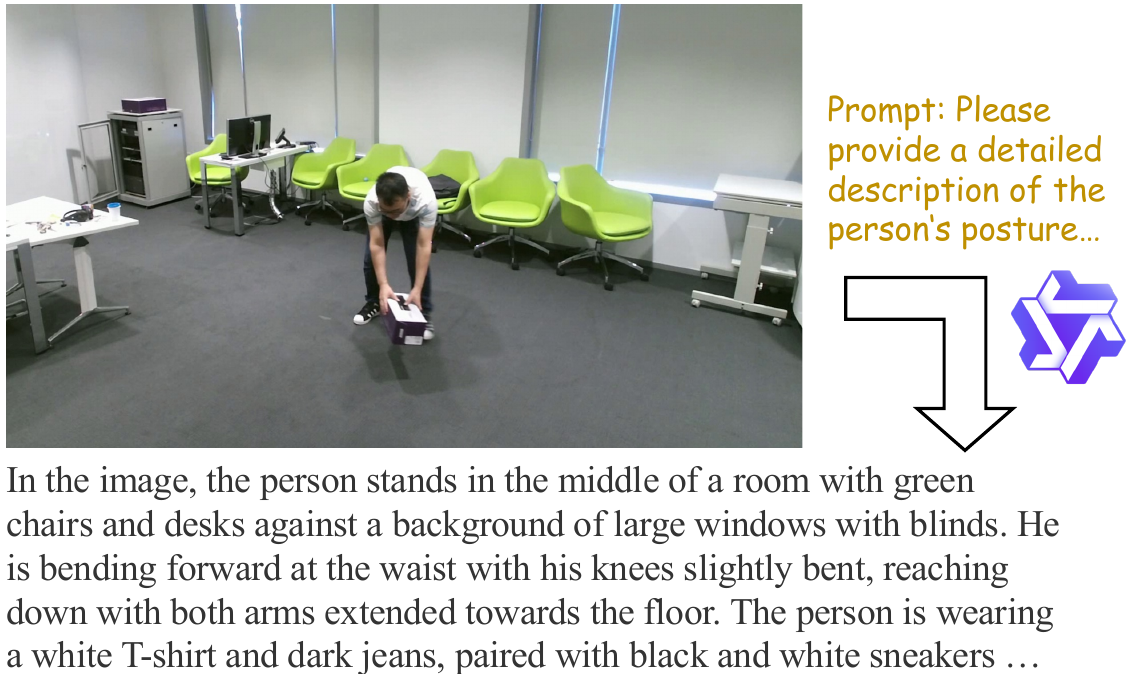}}
   \subfigure[Spatial differences mining task.]{
     \label{fig:SD}
       \includegraphics[width=\linewidth]{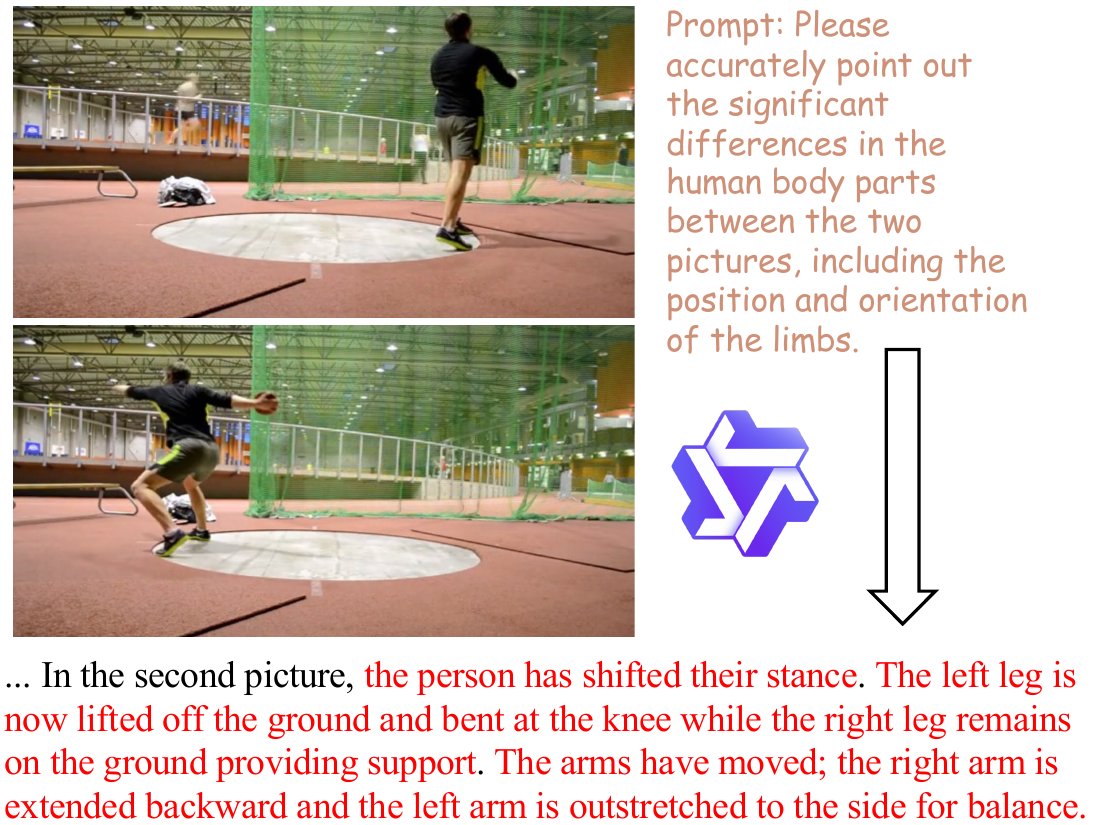}}
    \subfigure[Synthesized fine-grained QA]{
     \label{fig:FGQA}
       \includegraphics[width=\linewidth]{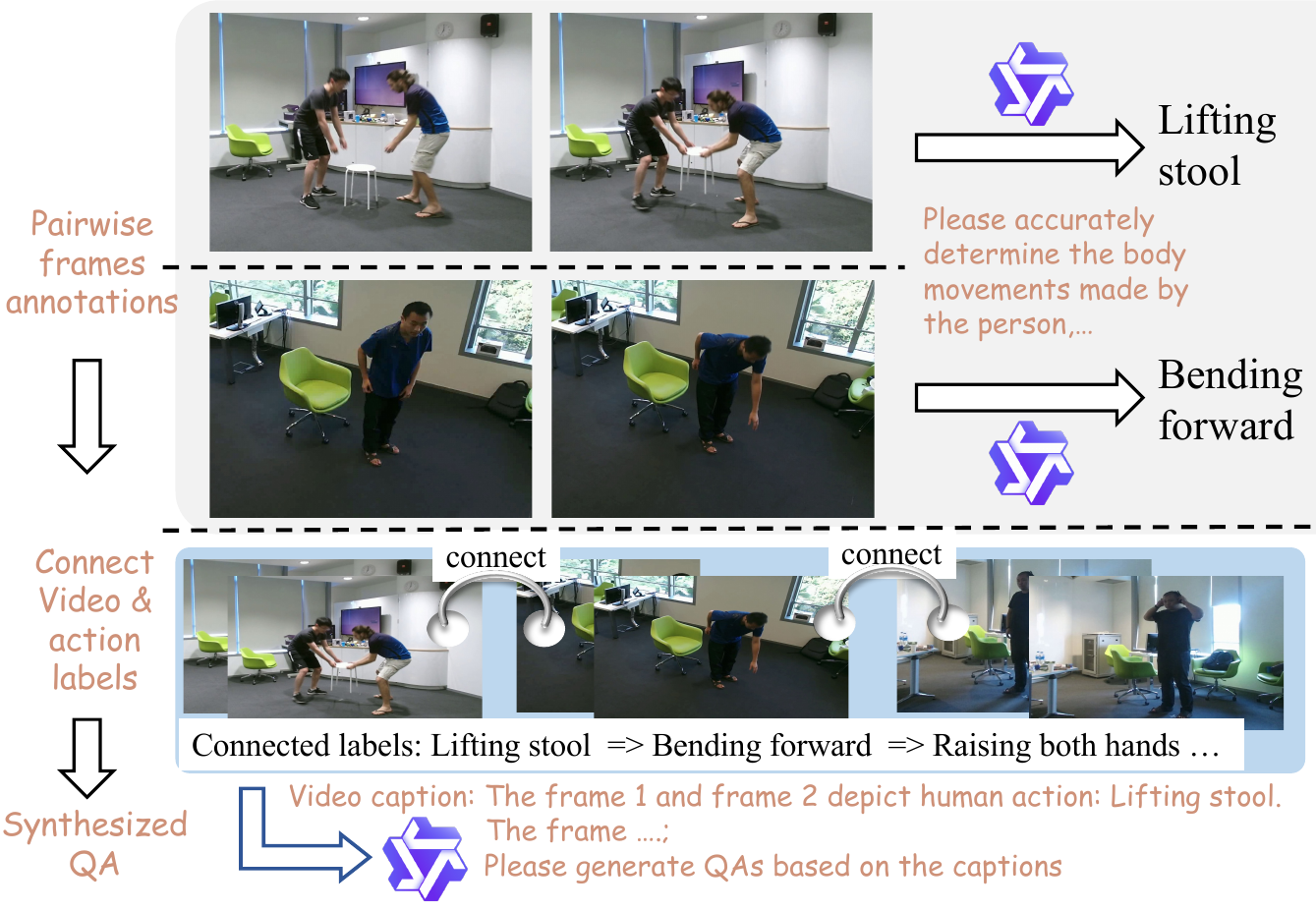}}
   \caption{Automated data collection pipelines for various proxy tasks. For pose description and spatial difference mining, we meticulously design prompts to guide advanced MLLMs in reasoning about human poses and the spatial differences between two images. To obtain fine-grained QA, we avoid directly generating QA from long videos, as it is challenging for MLLMs to comprehend them at a detailed granularity. Instead, we use MLLMs to annotate pairwise frames as action units and synthesize long videos through concatenation.}
    \label{fig:ablation}
\end{figure}

\subsection{Pose description}
Static spatial information understanding involves the analysis and interpretation of human pose information within each frame to improve the model's comprehension of static content.
We observe that existing MLLM could effectively describe the human pose in each frame. Hence, we can leverage advanced MLLM to automatically generate diverse labeled images for this task.

\begin{table*}[]
    \centering
    \begin{tabular}{|c|cccccccc|c|}
        \hline
        Model & AC & AR & AS & GS & HOI & LS & MV & TL & Avg. Acc. \\ \hline
        Human (10\%) & 71.7  & 93.5 & 78.8 & 94.8 & 92.5 & 93.3 & 80.0 & 79.4 & 87.4 \\ \hline
        $\text{GPT-4o}^{\dagger}$ & 32.5 & 71.4 & 51.1 & 68.3 & 66.3 & 56.3 & 47.1 & 30.4 & 54.7 \\ 
        $\text{Gemini}^{\dagger}$ & 27.8 & 62.1 & 50.5 & 74.8 & 65.2 & 49.0 & 48.5 & 36.5 & 53.9 \\ 
        \hline
        InternVL-2.5-78B & 54.0  & 61.4 & 55.2 & 82.9 & 73.0 & 60.8 & 48.2 & 53.5 & 62.2 \\
        Qwen2.5-VL-72B & 50.8 & 64.3 & 49.5 & 81.2 & 74.2 & 61.0 & 47.7 & 38.6 & 59.5 \\ \hline
        Oryx-7B & 39.7 & 64.3 & 52.4 & 78.6 & 75.3 & 58.8 & 44.6 & 33.1 & 57.3 \\ 
        Llava-ov-7B & 42.9 & 71.1 & 56.4 & 84.0 & 75.3 & 65.2 & 55.4 & 29.2 & 62.2 \\ 
        Llava-video-qwen2-7B & 42.9 & 69.8 & 58.6 & 84.5 & 78.7 & 68.8 & 46.8 & 37.7 & 63.2 \\
        InternVL-2.5-8B & 42.9 & 54.0 & 50.5 & 81.4 & 68.5 & 60.8 & 48.8 & \textbf{51.4} & 59.6 \\
        Qwen2.5-VL-7B & 41.3 & 59.8 & 44.8 & 78.6 & 70.8 & 55.3 & 44.9 & 30.4 & 54.7 \\
        $\text{PLLAVA-7B}^{\dagger}$ & 24.6 & 53.7 & 43.3 & 53.8 & 70.8 & 45.3 & 27.7 & 36.2 &44.1  \\
        Long-VITA & \textbf{46.8} & 56.9 & 50.2 & 77.9 & 67.4 & 55.3 & 45.4 & 40.1 & 56.3\\ 
        Cambrian-8B & 21.4 & 44.7 & 40.1 & 72.4 & 57.3 & 47.5 & 37.4 & 15.5 & 45.2  \\ \hline
        \textbf{Ours} & 45.2 & \textbf{82.6} & \textbf{61.8} & \textbf{85.5} & \textbf{83.1} & \textbf{75.7} & \textbf{59.6} & 42.5 & \textbf{69.4} \\ \hline
    \end{tabular}
    \vspace{-6pt}
    \caption{Comparison with advanced MLLMs on the ActionArt dataset. The highest performance among models with 7B parameters is highlighted in bold. Our method significantly outperforms all other open-source MLLMs, achieving the best results. By default, results are obtained by sampling 64 frames per video. Results marked with $\dagger$ correspond to those obtained with 16 frames per video.}
    \vspace{-6pt}
    \label{tab:sota}
\end{table*}

\subsection{Spatial differences mining}
Capturing changes in spatial information between adjacent frames is crucial for fine-grained video understanding. We develop a proxy task, called ``Spatial difference mining'', to train the model in capturing human pose transitions between adjacent frames.

Since we only need to understand the posture differences between two frames, we can input both images fully, without compression, into a Multimodal Large Language Model (MLLM) without worrying about exceeding the predefined token length. With appropriate prompting, existing MLLMs can effectively comprehend the changes in human posture between the two images, even when the images are uncompressed and limited to just two. Therefore, we can leverage existing multimodal large models to automatically generate the data required for this proxy task.

By leveraging automatically generated datasets, we can enable the continuous tracking of motion trajectories or pose changes, thus improving the model's understanding of action transitions.

\subsection{Synthesized fine-grained QA}
Temporal content understanding requires the model to grasp the sequence of actions throughout an entire video. Typically, long videos contain rich temporal variations.
However, obtaining fine-grained captions for long videos is challenging and prone to noise. 

To address this, we employ a concatenation method to create long videos and form action sequences in fine granularity.
Specifically, since two frames can depict an human action, we consider two frames from a video as an action unit.
These action units are fed into existing MLLM to assign fine-grained action labels, such as ``raise right hand'' or ``bend over''.
After obtaining the action units and the corresponding labels, we concatenate different action units and their labels to form a pseudo long video with fine-grained action sequence labels.
Notably, the concatenated action units could from different videos, which makes our generation pipeline more scalable.
We prompting the advanced MLLM to generate diverse QA according to the action sequence for each video.

By integrating these methods, we aim for the model to gradually improve its ability to understand fine-grained actions without relying on large-scale manually annotated data. The generation of auxiliary task data can be facilitated through automated tools, reducing manual annotation costs and improving training efficiency.

\section{Experiments}

In this section, we evaluate advanced MLLM in our ActionArt, and report the results.
And then, we conduct experiments to demonstrate the effectiveness of our proposed proxy tasks.

\subsection{Dataset and evaluation metric}
There are 2,678 manual annotated multi-choice QAs in our ActionArt, which are further divided into seven tasks.
We utilize the accuracy in each sub-tasks and the average accuracy in the whole benchmark as our evaluation metric.

\subsection{Implementation details}
\label{sec:exp_id}

We use the Oryx-7B-Image~\cite{oryx} model as our baseline. During training, we constrain the maximum number of sampled frames to 64, with each frame's resolution capped at 480x480 pixels.
We extract visual tokens from each frame and organize these tokens in a SlowFast~\cite{sf_llava,slowfast} manner to derive the video representation. 
Our training process comprises three stages. First, we train the model on the Oryx-Video-SFT dataset~\cite{oryx} to adapt to the SlowFast approach for video representation, rather than simply concatenating visual tokens from each frame. In this setup, the spatial pooling stride is set to 2 in the slow branch and 4 in the fast branch.
Next, we train the model using our proxy tasks.

In the final stage, the model is trained on our manually annotated dataset. At this point, only the manually annotated captions are utilized. We employ the advanced language model GPT-4~\cite{openai2023gpt4} to generate diverse QA pairs from these captions for training purposes.

\noindent \textbf{Data collection of proxy tasks.}
Our proxy tasks utilize data from the NTU-RGB+D-120 dataset~\cite{nturgbd} and the Kinetics700 (abbreviated as K700) dataset~\cite{K700}. The NTU-RGB+D-120 dataset features a wide range of human actions with high video quality, while the K700 dataset encompasses various human actions across different domains.

Given the distinct characteristics of these datasets, we address the pose description task by prompting Qwen2.5-VL-72B~\cite{qwen2.5} to describe the human poses in 50K randomly selected frames from the NTU-RGB+D-120 dataset. For the spatial difference mining task, we task Qwen2.5-VL-72B~\cite{qwen2.5} to identify pose differences between two evenly sampled frames in the K700 videos, ultimately selecting 200K frames for training.

For fine-grained QA, we employ Qwen2.5-VL-72B~\cite{qwen2.5} to infer the detailed actions depicted in two frames and create a labeled action unit set. We then concatenate multiple action units to synthesize a longer video, prompting Qwen-2.5~\cite{qwen2.5} with the action unit labels to generate a variety of QA pairs. In total, we use 200K synthesized videos for training.

\begin{table*}[h]
    \centering
    \begin{tabular}{|c|ccc|c|cccccccc|c|}
        \hline
    \multirow{2}{*}{\color{blue}PF}  &  \multicolumn{4}{|c|}{Task} & \multirow{2}{*}{AC} & \multirow{2}{*}{AR} & \multirow{2}{*}{AS} & \multirow{2}{*}{GS} & \multirow{2}{*}{HOI} & \multirow{2}{*}{LS} & \multirow{2}{*}{MV} & \multirow{2}{*}{TL} & \multirow{2}{*}{Avg. Acc.} \\ \cline{2-5}
      &   PD & SD & FGQA & MA & \multicolumn{8}{c|}{} &  \\ \hline
   \XSolidBrush   &   \XSolidBrush & \XSolidBrush & \XSolidBrush & \XSolidBrush & 40.5 & 58.5 & 50.8 & 81.2 & 74.2 & 54.3 & 44.0 & 32.8 & 55.8 \\ \cdashline{1-14}
        \multicolumn{14}{|c|}{Stage 2. Proxy Tasks Learning}  \\
        \cdashline{1-14}
   \XSolidBrush &   \XSolidBrush & \XSolidBrush & \CheckmarkBold & \XSolidBrush & 46.0 & 66.9 & 54.2 & 79.9 & 77.5 & 62.2 & 47.1 & 34.0 & 59.6 {\color{red}(+3.8)} \\ 
  \rowcolor{cyan!40}
  \CheckmarkBold &   \XSolidBrush & \XSolidBrush & \CheckmarkBold & \XSolidBrush & 48.4 & 72.7 & 57.3 & 82.3 & 75.3 & 64.3 & 45.4 & 33.4 & 61.4 {\color{red}(+5.6)} \\ 
    
    \XSolidBrush &    \CheckmarkBold & \XSolidBrush & \CheckmarkBold & \XSolidBrush & 50.0 & 72.1 & 56.1 & 82.5 & 73.0 & 62.0 & 44.9 & 31.6 & 60.4 {\color{red}(+4.6)} \\ 
    \rowcolor{cyan!40}
    \CheckmarkBold &  \CheckmarkBold & \XSolidBrush & \CheckmarkBold & \XSolidBrush & 46.8 & 75.2 & 55.5 & 84.0 & 73.0 & 64.5 & 46.3 & 31.6 & 61.6 {\color{red}(+5.8)} \\
        
    \XSolidBrush  &    \XSolidBrush & \CheckmarkBold & \CheckmarkBold & \XSolidBrush & 46.0 & 71.4 & 55.8 & 82.3 & 75.3 & 66.3 & 44.0 & 32.8 & 61.1 {\color{red}(+5.3)} \\
    \rowcolor{cyan!40}
    
    \CheckmarkBold &    \XSolidBrush & \CheckmarkBold & \CheckmarkBold & \XSolidBrush & 48.4 & 73.0 & 58.0 & 84.2 & 76.4 & 67.8 & 47.9 & 33.7 & 63.0 {\color{red}(+7.2)} \\
   
    \XSolidBrush &  \CheckmarkBold & \CheckmarkBold & \CheckmarkBold & \XSolidBrush & 50.0 & 72.3 & 56.7 & 81.2 & 73.0 & 64.7 & 45.4 & 31.3 & 60.9 {\color{red}(+5.1)} \\
\rowcolor{cyan!40}
    \CheckmarkBold &  \CheckmarkBold & \CheckmarkBold & \CheckmarkBold & \XSolidBrush & 46.0 & 72.7 & 59.9 & 83.6 & 78.7 & 67.7 & 47.4 & 31.9 & 62.8 {\color{red}(+7.0)} \\
    
        \cdashline{1-14}
        \multicolumn{14}{|c|}{Stage 3. Training with manual annotations}  \\
        \cdashline{1-14}
        \rowcolor{gray!40}  
   \XSolidBrush &  \XSolidBrush & \XSolidBrush & \XSolidBrush & \CheckmarkBold & 39.7 & 82.6 & 60.8 & 84.9 & 79.8 & 70.5 & 54.0 & 48.6 & 67.6 {\color{red}(+11.8)} \\ 
        
   \XSolidBrush & \CheckmarkBold & \CheckmarkBold & \CheckmarkBold & \CheckmarkBold & 45.2 & 82.6 & 61.8 & 85.5 & 83.1 & 75.7 & 59.6 & 42.5 & 69.4 {\color{red}(+13.6)} \\ \hline
        
    \end{tabular}

    \caption{Ablation studies in ActionArt dataset. ``PD'' denotes pose description task, ``SD'' denotes the spatial differences mining task and ``FGQA'' denotes the fine-grained QA.
    ``MA'' means using manual annotations.
    To better demonstrate the effectiveness of the proposed proxy tasks, we report the results of directly training with the manual labeled data, marked in \colorbox{gray!40}{gray}.
    The improvement over the baseline model is marked in {\color{red} red}. 
    {\color{blue}``PF''} means the parameter fusion strategy. Please refer text for details.}
    \label{tab:ab}

\end{table*}

\subsection{Main results in ActionArt}
We compared our model with the advanced MLLM including proprietary MLLM: GPT-4o~\cite{openai2024gpt4o}, Gemini~\cite{geminiteam2024geminifamilyhighlycapable}; and open-sourced MLLM including InternVL-2.5~\cite{wang2024internvideo2}, Qwen2.5-VL~\cite{qwen2.5}, Oryx-7B~\cite{oryx}, Llava-onevision-7B~\cite{li2024llavaone}, Llava-video-qwen2-7B~\cite{zhang2024llavanextvideo}, PLLAVA-7B~\cite{xu2024pllava}, Long-VITA~\cite{shen2025longvita} and Cambrian-8B~\cite{tong2024cambrian}.
Apart from the MLLM, we sampled 10\% of the questions in each task and had 5 volunteers answer them to calculate the human accuracy.
The experimental results are shown in Table~\ref{tab:sota}.

From the Table~\ref{tab:sota}, we can observe that there is significant room for improvement in the performance of current multimodal large language models on the ActionArt dataset. For instance, Qwen2.5-VL-72B, despite its large parameter size, achieves an accuracy of only 59.5\%. Different models exhibit varying strengths across different subtasks. Llava-ov-7B performs well in action recognition, local spatial understanding, and moving direction tasks, while InternVL-2.5-8B excels in temporal localization, outperforming other models significantly. These results suggest that current multimodal large models still lack sufficient granularity in human-centric video understanding.

Furthermore, the results indicate that achieving fine-grained, human-centric understanding is challenging, as simply increasing the model's parameters results in only marginal improvements. For example, scaling the Qwen-2.5-VL model from 7B to 72B parameters yields only a 4.8\% performance gain.

In contrast, our proposed model demonstrates superior performance compared to existing methods, highlighting the efficacy and value of our approach. We achieve an overall accuracy of 69.4\% on the benchmark, which is 7.2\% higher than the second-best model with 7B parameters and surpasses the larger Qwen-2.5-VL-72B model. Our model delivers the best performance across all sub-tasks, except for the action count task, where it falls short by 1.6\% compared to the top 7B model, Long-VITA. This shortcoming is attributed to our focus on designing proxy tasks for human-centric video understanding, which does not address the inherent limitations in existing MLLM architectures. Action counting requires dense frame sampling for accurate computation, but the limited context length of current MLLMs restricts their ability to sample frames densely. Unlike our approach, Long-VITA emphasizes long video understanding and employs an advanced sampling strategy, enabling it to achieve superior results in this area.

\subsection{Ablation study}

We conducted extensive experiments to showcase the effectiveness of our proposed proxy tasks. The experimental results are presented in Table~\ref{tab:ab}. The first row displays the results for our baseline, which is the Oryx-7B model without any fine-tuning on proxy tasks or manually labeled data.

The training data used in our proxy tasks does not consist of actual video sequences. For instance, the data employed for spatial differences mining comprises pairwise frames. Consequently, we speculate that these proxy tasks might weaken the model's capability to handle long sequences. To address this challenge, we further explore a simple solution: parameter fusion strategy.  

\noindent \textbf{- Effectiveness of Fine-Grained QA (FGQA).}
Training the model with the proposed fine-grained QAs, automatically synthesized from existing MultiLingual Language Models (MLLM), noticeably enhances model performance across tasks. The average performance increases from 55.8\% to 59.6\%, marking a 3.8\% improvement over the baseline. These experiments demonstrate that our fine-grained QA approach effectively guides the model to understand human-centric videos at a more detailed level, reducing the requirement for manual annotations.

\begin{figure}[tp]
    \centering
    \includegraphics[width=\linewidth]{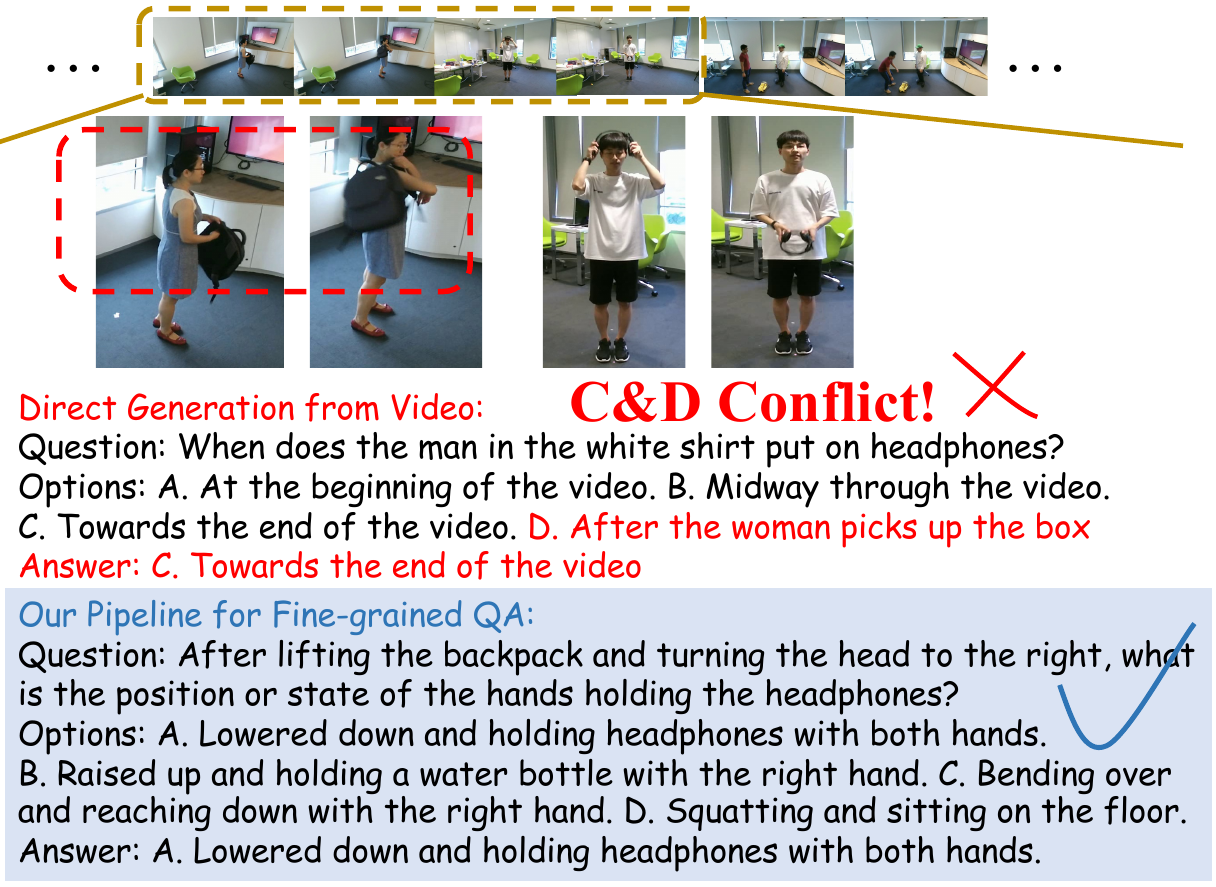}
    \caption{Comparison of direct QA generation from videos and our proposed pipeline. Directly generating QA from videos often lead to conflicts. As in the figure, both options C \& D is correct for the question ``When does the man put on the headphones?''. As the granularity of the QA increases, conflicts are more likely to arise because existing MLLMs struggle to perceive all the fine-grained details in a video.}
    \label{fig:QA_comp}
\end{figure}

The performance improvement stems from our method of generating fine-grained QA. Instead of prompting the MLLM to generate QA from an entire video, we prompt it to infer actions between two frames. By concatenating these pairwise frames, we synthesize a longer video and derive fine-grained labels, enhancing the model's understanding.
Fig.~\ref{fig:QA_comp} illustrates our generated QA, highlighting the effectiveness of this approach.

\noindent \textbf{- Effectiveness of the Pose Description (PD) Task.}
As observed in Table~\ref{tab:ab}, the pose description (PD) task significantly enhances the model's ability to recognize human actions and understand spatial information. This likely occurs because both action recognition and spatial understanding require accurate perception of spatial information within videos. Training the model with both PD and FGQA tasks elevates average accuracy from 55.8\% to 60.4\%, demonstrating a performance gain of 4.6\% over the baseline.

\noindent \textbf{- Effectiveness of the Spatial Difference (SD) Task.}
Compared to PD, the SD task is more challenging as it requires understanding spatial information changes between two frames. The combination of the SD and FGQA tasks shows improved performance, with average accuracy rising to 61.1\%, 5.3\% higher than the baseline. However, including PD alongside SD and FGQA only achieves a limited gain, with average accuracy reaching 60.9\%. This may be because the knowledge acquired through PD is partly encompassed within the SD task, which also requires understanding pose information.

Nevertheless, the proposed proxy tasks notably improve model performance, reaching 61.1\% average accuracy. For reference, training the model using manual annotations, as described in Sec.~\ref{sec:exp_id}, results in a 67.6\% average accuracy, showcasing the proxy tasks' ability to reduce reliance on costly manual labels.

\noindent \textbf{- Combining Proxy Tasks and Manual Annotations.}
Our proxy tasks, when combined with manual annotations, further enhance performance. As detailed in Sec.~\ref{sec:exp_id}, our final model is initially trained on the proxy tasks before using manually annotated data. The last row in Table~\ref{tab:ab} shows that, compared to training solely with manual annotations, pretraining on proxy tasks boosts model accuracy from 67.6\% to 69.4\%.
Proxy tasks improve performance across most tasks, except for temporal localization. This might be due to the lack of real long videos in the proxy task training, affecting the model's ability to handle temporal changes.

\noindent \textbf{- Parameter fusion strategy.}
Based on the experimental analysis above, proxy tasks may weaken the model's ability to capture long sequences. 
Therefore, we investigate a parameter fusion strategy for this issue: after training with the proxy tasks, we apply a weighted average of the model parameters with the baseline model, which is trained on video instruction tuning dataset.
The results are shown in Table~\ref{tab:ab}. We can observe that, with the parameter fusion strategy, the proposed proxy tasks can further improve the model performance to 63.0\%, which is only 4.6\% lower than the model merely trained with manual labeled data.

\section{Conclusion}
In this work, we present ActionArt, a challenging benchmark crafted to evaluate the performance of MLLMs in fine-grained human-centric video understanding. We break down this task into seven sub-tasks to assess the models' capabilities across various dimensions. Our experimental results reveal that current MLLMs face difficulties in capturing fine-grained spatial details and temporal variations in human-centric videos.
Due to the high cost associated with obtaining manual labels, we examine the inherent challenges of fine-grained human-centric video understanding and devise corresponding proxy tasks as a solution. Through the careful design of these tasks, we facilitate the automatic generation of training data, thereby reducing dependence on expensive manual labeling. Extensive experiments show that our proposed proxy tasks significantly enhance model performance.

\noindent \textbf{Limitations and future work. } Our work is developed from a data-driven perspective, and as such, it is constrained by the existing limitations of current MLLMs. For example, the restricted context length limits the number of input frames, while detailed tasks such as counting human actions require more frames to be effective. However, our approach holds potential for future integration with MLLM architecture optimization efforts to overcome these constrains.

{
    \small
    \bibliographystyle{ieeenat_fullname}
    \bibliography{main}
}

\end{document}